%% file: main.tex
\pdfoutput=1
\documentclass[10pt,twocolumn,letterpaper]{article}

\usepackage[pagenumbers]{cvpr} % To force page numbers, e.g. for an arXiv version

\usepackage{graphicx}
\usepackage{amsmath}
\usepackage{amssymb}
\usepackage{booktabs}
\usepackage{lipsum}

\input{shortcuts.tex}

\usepackage[capitalize]{cleveref}
\crefname{section}{Sec.}{Secs.}
\Crefname{section}{Section}{Sections}
\Crefname{table}{Table}{Tables}
\crefname{table}{Tab.}{Tabs.}

\begin{document}

\title{SAM3D: Segment Anything in 3D Scenes}

\author{
Yunhan Yang$^1$ \authorskip Xiaoyang Wu$^2$ \authorskip Tong He$^1$ \authorskip Hengshuang Zhao$^2$ \authorskip Xihui Liu$^2$ \\
$^1$Shanghai Artificial Intelligence Lab \authorskip $^2$The University of Hong Kong\\
{\tt\small \url{https://github.com/Pointcept/SegmentAnything3D}}
}
\twocolumn[{%
\renewcommand\twocolumn[1][]{#1}%
\maketitle
\begin{center}
\centering
\includegraphics [width=0.95\textwidth]{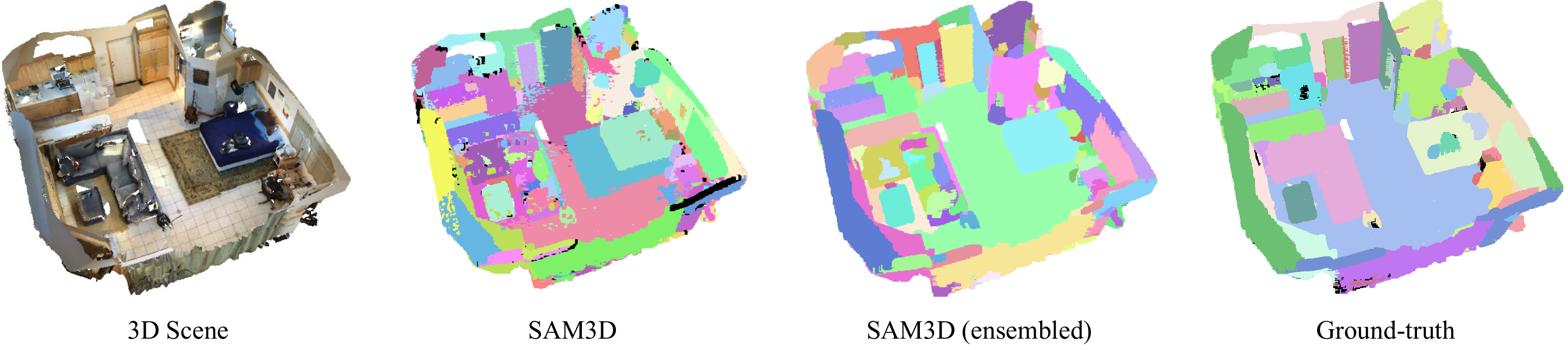}
\captionof{figure}{\textbf{Qualitative results of our SAM3D.} The first subfigure demonstrate the 3D scene input. The second subfigure is the segmentation masks predicted by SAM3D. The third subfigure is refined masks generated by ensembling the SAM result and the over-segmentation result. The last subfigure is the ground-truth segmentation labels in ScanNet~\cite{dai2017scannet}. }
\label{fig:fig1}
\end{center}%
}]

%%%%%%%%% ABSTRACT %%%%%%%%%
\begin{abstract}
  \input{./tex/0_abstract}
\end{abstract}

%%%%%%%%% BODY %%%%%%%%%
\input{./tex/1_introduction}

\input{./tex/2_related}

\input{./tex/3_method}

\input{./tex/4_experiments}

\input{./tex/5_conclusion}

%%%%%%%%% REFERENCE %%%%%%%%%
{
% \clearpage
\small
\bibliographystyle{ieee_fullname}
\bibliography{main}
}

%%%%%%%%% APPENDIX %%%%%%%%%
% \clearpage
% \newpage
% \appendix
% \section*{Appendix}
% \input{./tex/6_appendix}

\end{document}

%% file: shortcuts.tex
\usepackage{array}
\usepackage{times}
\usepackage{epsfig}
\usepackage{graphicx}
\usepackage{float}
\usepackage{wrapfig}
\usepackage{amsmath,amssymb,amsthm}
\usepackage{algorithm,algorithmicx,algpseudocode}
\usepackage{bm,xspace}
\usepackage{comment}
\usepackage{multirow}
\usepackage{balance}
\usepackage{url}
\usepackage{booktabs}
\usepackage{etoolbox,siunitx}
\usepackage{calc}
\usepackage{pifont,hologo}
\usepackage{color}
\usepackage{adjustbox}
\usepackage[normalem]{ulem}  % \xout
\usepackage[table]{xcolor}
\usepackage[pagebackref,breaklinks,colorlinks,linkcolor=red,citecolor=blue]{hyperref}
\usepackage[accsupp]{axessibility}  % Improves PDF readability for those with disabilities.

\newcommand{\authorskip}{\hspace{4mm}}

\definecolor{blue}{HTML}{0055cc}
\definecolor{red}{HTML}{cc1100}
\definecolor{orange}{HTML}{cc7700}
\definecolor{gray}{HTML}{efefef}
\definecolor{darkgreen}{rgb}{0.13, 0.55, 0.13}
\definecolor{darkgray}{HTML}{757575}

\newcommand{\figref}[1]{Figure~\ref{#1}}

\renewcommand{\eqref}[1]{Eq.~\ref{#1}}

\newcolumntype{x}[1]{>{\centering\arraybackslash}p{#1}}
\newcolumntype{y}[1]{>{\raggedright\arraybackslash}p{#1}}
\newcolumntype{z}[1]{>{\raggedleft\arraybackslash}p{#1}}

\DeclareMathSymbol{@}{\mathord}{letters}{"3B}

\makeatletter
\DeclareRobustCommand\onedot{\futurelet\@let@token\@onedot}
\def\@onedot{\ifx\@let@token.\else.\null\fi\xspace}

\newcommand*{\Rom}[1]{\expandafter\@slowromancap\romannumeral #1@}
\newcommand*{\rom}[1]{\expandafter\romannumeral #1}

\def\ceil#1{\lceil #1 \rceil}

\def\1{\bm{1}}

\DeclareMathAlphabet{\mathsfit}{\encodingdefault}{\sfdefault}{m}{sl}
\SetMathAlphabet{\mathsfit}{bold}{\encodingdefault}{\sfdefault}{bx}{n}

%% file: tex/0_abstract.tex
In this work, we propose SAM3D, a novel framework that is able to predict masks in 3D point clouds by leveraging the Segment-Anything Model (SAM) in RGB images without further training or finetuning.
For a point cloud of a 3D scene with posed RGB images, we first predict segmentation masks of RGB images with SAM, and then project the 2D masks into the 3D points.
Later, we merge the 3D masks iteratively with a bottom-up merging approach. At each step, we merge the point cloud masks of two adjacent frames with the bidirectional merging approach. In this way, the 3D masks predicted from different frames are gradually merged into the 3D masks of the whole 3D scene. Finally, we can optionally ensemble the result from our SAM3D with the over-segmentation results based on the geometric information of the 3D scenes.
Our approach is experimented with ScanNet dataset and qualitative results demonstrate that our SAM3D achieves reasonable and fine-grained 3D segmentation results without any training or finetuning of SAM.

%% file: tex/1_introduction.tex
\begin{figure*}[!t]
    \centering
    \includegraphics[width=\linewidth]{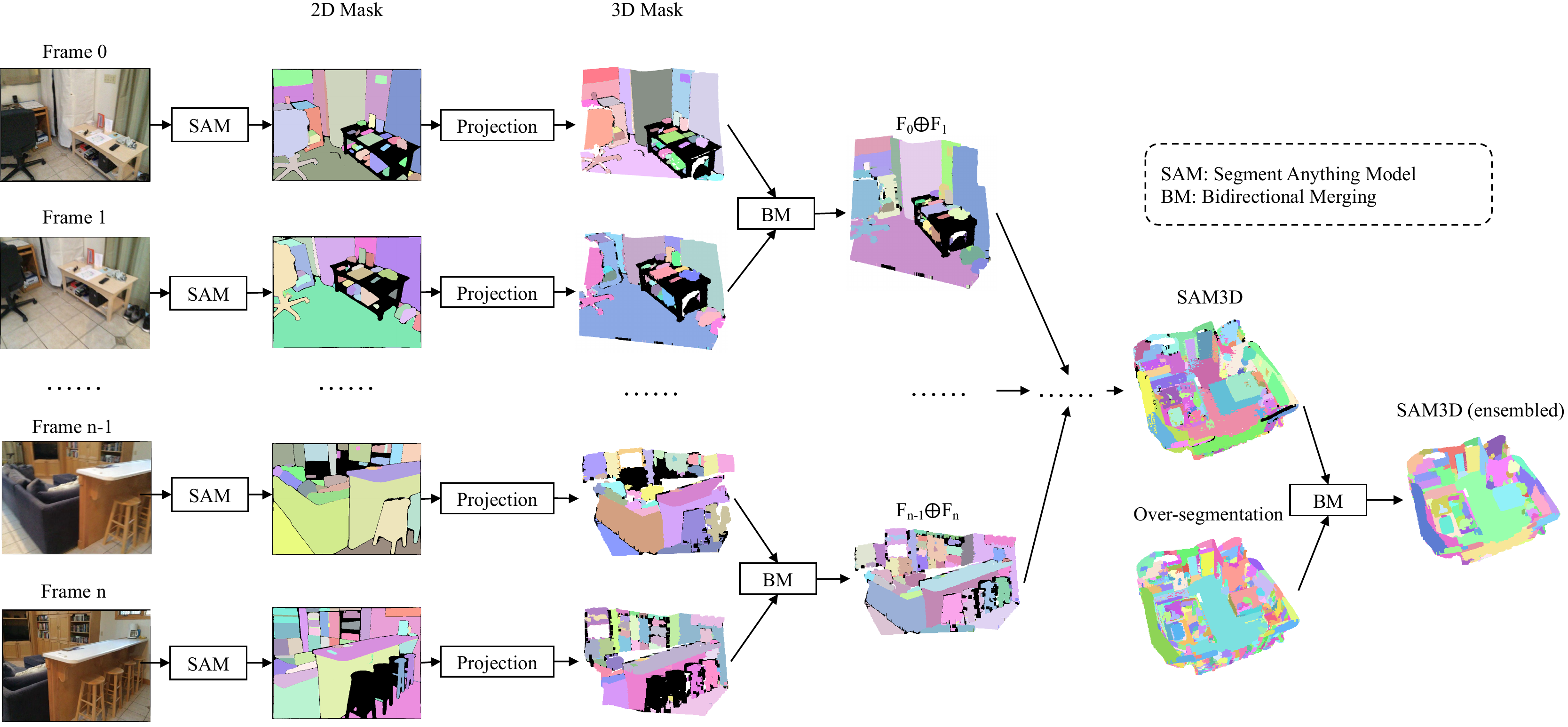}
    \caption{\textbf{Overview.} For input images, we first use SAM to generate 2D masks, and then map the 2D masks to 3D. Then we iteratively merge adjacent point clouds with the Bidirectional Merging (BM) approach until we obtain the 3D masks of the whole scene. We finally merge the SAM3D result with the over-segmentation masks to obtain an ensembled result.}
    \label{fig:pipeline}
\end{figure*}

\begin{figure*}[!t]
    \centering
    \includegraphics[width=\linewidth]{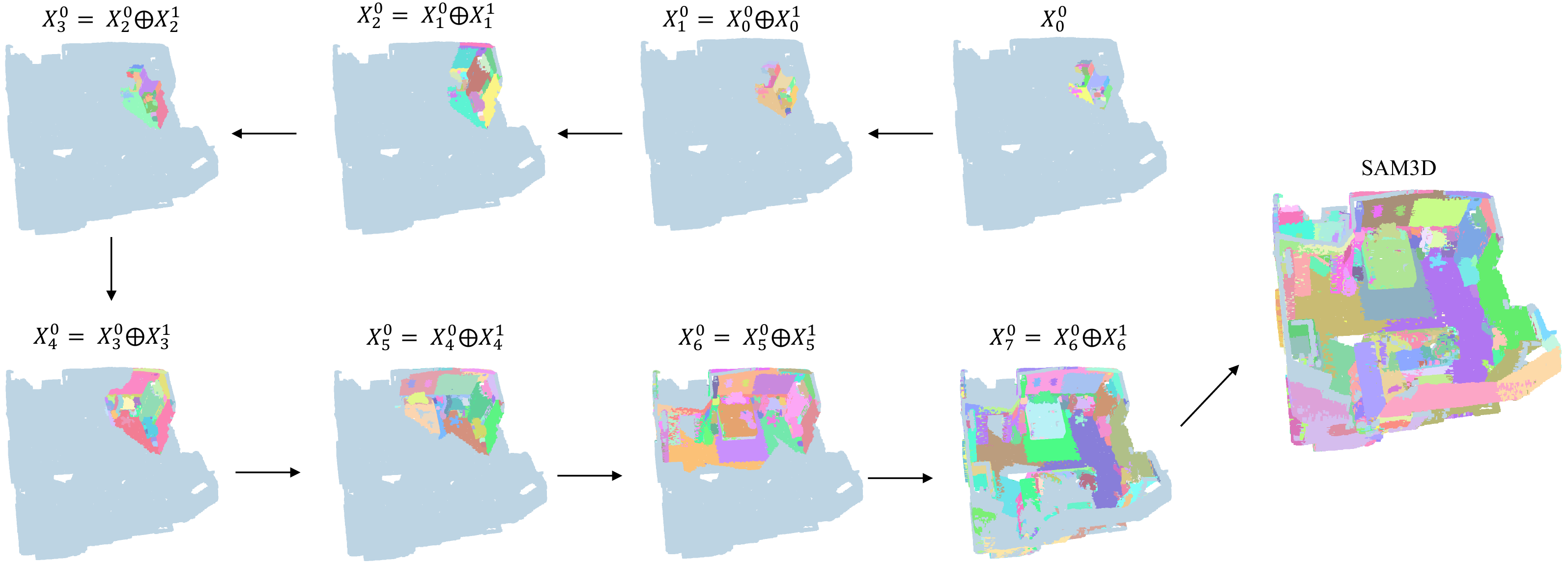}
    \caption{\textbf{Bottom-up Merging.} We illustrate the process of gradually expanding the partial point cloud masks with the bottom-up merging approach.}
    \label{fig:rmm}
\end{figure*}

\section{Introduction and Background}

3D segmentation aims to predict point-level semantic labels of the 3D point cloud of a scene. Previous approaches either develop semantic segmentation approaches on point cloud~\cite{qi2017pointnet, qi2017pointnetplusplus, li2018pointcnn, wu2018pointconv}, or leverage features extracted from RGB images by cross-modal fusion between 2D and 3D~\cite{dai20183dmv, jaritz2019multiview, kundu2020virtual}.

While 3D point clouds provide geometric information that cannot be obtained from images, the 2D visual perception models have demonstrated great success because of the large-scale RGB images and corresponding annotations as training data.
Recently, the Segment Anything Model (SAM)~\cite{kirillov2023segany} has been introduced for promptable, fine-grained segmentation of RGB images.
SAM has achieved astonishing results in fine-grained segmentation of stuff, objects, and object parts in RGB images of complex scenes.
Follow-up works SAD~\cite{cen2023sad} leverages SAM to segment the depth maps. With the cues with enhanced geometry information, it addresses the over-segmentation problem by SAM.
Another work Anything-3D~\cite{shen2023anything3d} combines SAM with BLIP~\cite{li2022blip}, Stable Diffusion~\cite{rombach2021highresolution}, and Point-E~\cite{nichol2022pointe} for single-view conditioned 3D reconstruction of objects in an image.

In this project, we investigate how to leverage the segmentation results of SAM to generate fine-grained 3D masks for 3D scenes.
We propose SAM3D, which projects the SAM segmentation masks from RGB frames into the 3D scene, and merge the results iteratively to get the segmentation masks of the whole 3D scene.
Specifically, given the point cloud of a 3D scene with posed RGB images, we first predict 2D segmentation masks by applying SAM on the RGB images, and then project the 2D masks into 3D. Then the 3D masks of partial scenes are merged iteratively in a bottom-up way to generate the 3D masks of the whole scene. We propose a bidirectional merging approach to merge the two masks from adjacent frames at each merging step.
Finally, we ensemble the 3D masks with the over-segmentation masks of the scene.
Our proposed approach takes advantage of SAM and does not require training or finetuning the model.
We demonstrate the qualitative results on 3D scenes of ScanNet~\cite{dai2017scannet} dataset.

%% file: tex/2_related.tex
% \section{Related Work}
% 222

%% file: tex/3_method.tex
\begin{figure*}
    \centering
    \includegraphics[width=\linewidth]{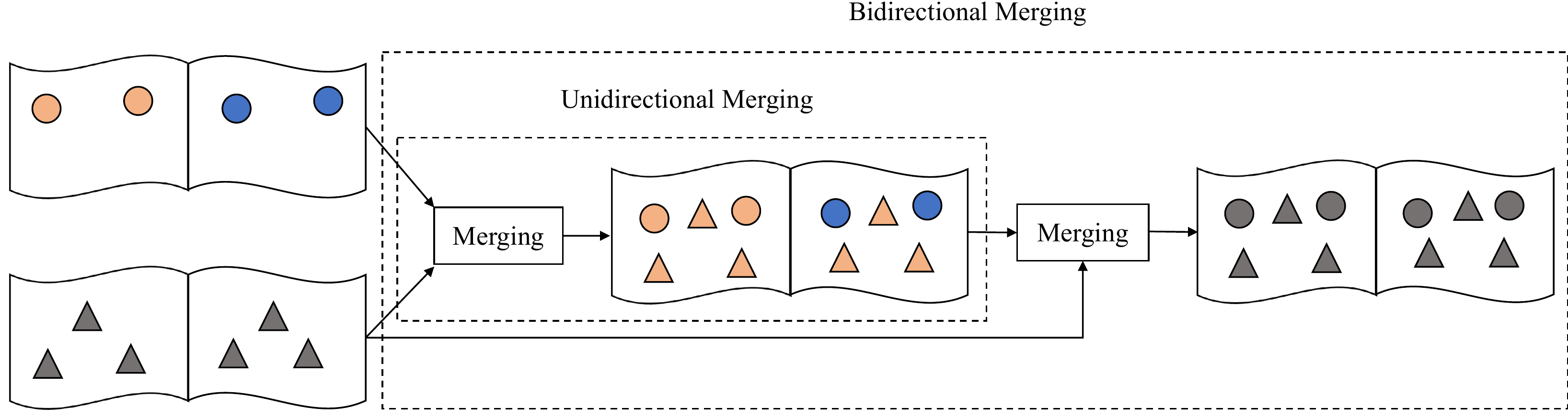}
    \caption{
    \textbf{Illustration of the Bidirectional Merging approach.} We use different colors to denote different mask ids. Our bidirectional merging approach merges masks from different frames to obtain unified masks in the scene.}
    \label{fig:bmm}
\end{figure*}
\begin{figure*}[!t]
    \centering
    \includegraphics[width=\linewidth]{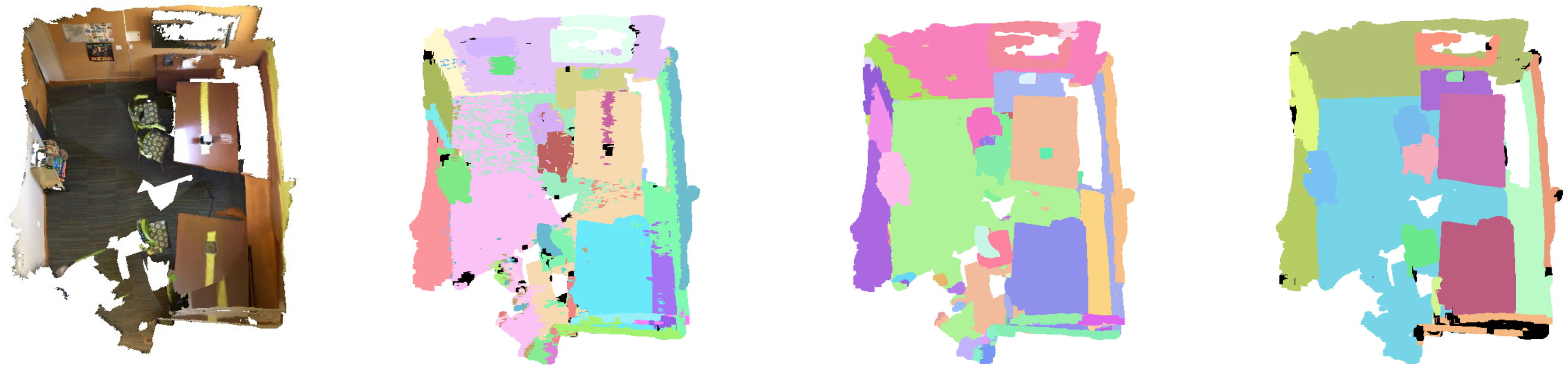}
    \vspace{1em}
    
    \includegraphics[width=\linewidth]{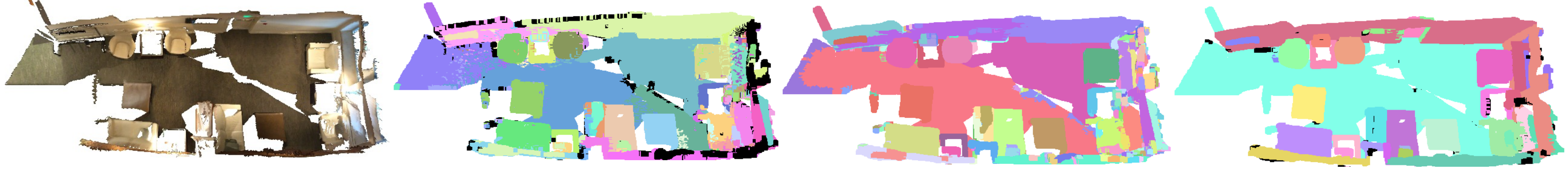}
    \vspace{1em}
    
    \includegraphics[width=\linewidth]{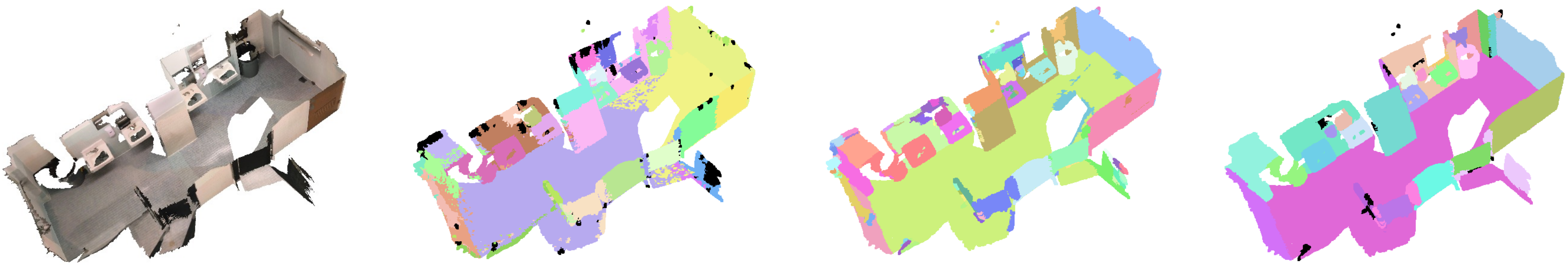}
    \vspace{1em}
    
    \includegraphics[width=0.95\linewidth]{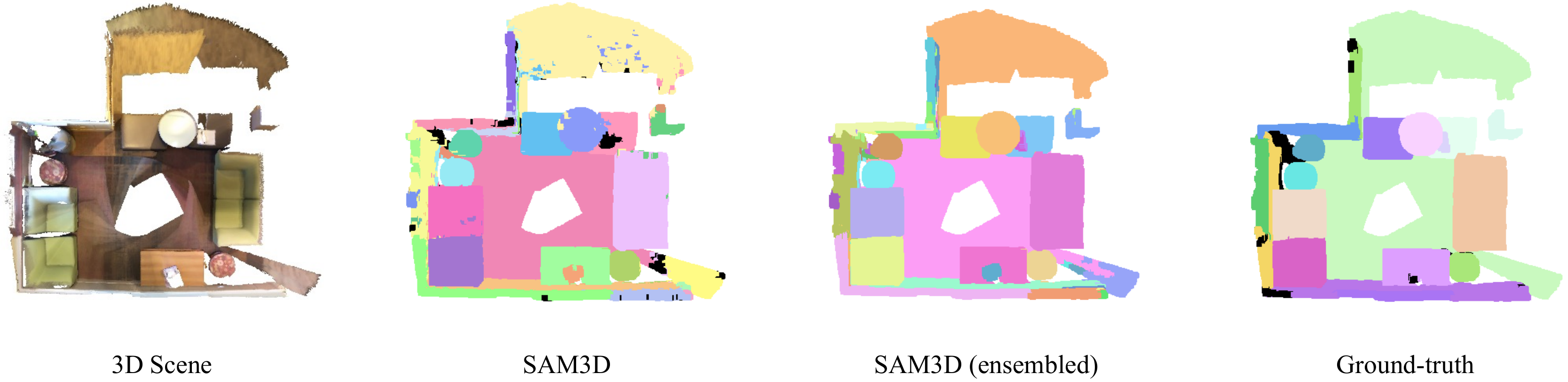}
    \caption{\textbf{Qualitative Segmentation Results.} Our approach generates high-quality instance masks at multiple scales. Different color represents group index only.}
    \label{fig:results}
\end{figure*}
\section{Method}
We propose an approach to lift the SAM~\cite{kirillov2023segany} segmentation results from the posed RGB images to point clouds, as shown in~\figref{fig:pipeline}. 
Given a point cloud with a set of posed RGB images, we first map the SAM segmentation results of RGB frames to corresponding point cloud frames.
Then we propose a bidirectional merging approach to merge each pair of 3D segmentation results from adjacent frames.
We iteratively merge the 3D masks in a bottom-up way to obtain the 3D masks of the whole scene.
Finally, we ensemble the SAM-predicted results with the over-segmentation~\cite{dai2017scannet} results to obtain finer segmentation maps.
\subsection{Single-frame 3D masks from SAM}
\label{sec:sgm}

We first apply SAM~\cite{kirillov2023segany} automatic mask generation approach on the single-frame RGB images to obtain the pixel-level masks of an image.
SAM returns nested masks with different granularity: whole, part, and subpart.
In order to obtain non-overlapping masks, if a pixel is covered by multiple masks, we assign the mask ID with the highest predicted IoU to the pixel.
Then we map the 2D masks to the 3D space according to the depth of each pixel provided by the RGB-D image. Specifically, the mapping is formulated as,
\begin{align}
[x_{i} ,y_{i} ,z_{i} ]^{T} =\mathcal{R}^{-1}\mathcal{M}^{-1}s[u_{i} ,v_{i} ,1]^{T}-\mathcal{R}^{-1}T,
\end{align}
where $[u_{i},v_{i},1]$ and $[x_{i},y_{i},z_{i}]$ are 2D homogeneous coordinates and projected 3D homogeneous coordinates, respectively.
$\mathcal{M}$ is the intrinsic camera calibration matrix.
$\mathcal{R}$ and $\mathcal{T}$ are the rotation matrix and translation matrix obtained from the extrinsic camera pose matrix, and s is a scaling factor.
Finally, we apply grid pooling~\cite{wu2022point} to downsample the point cloud masks.

\subsection{Bidirectional Merging between Two Point Clouds}
\label{sec:mta}

Given the 3D masks derived from the previous subsection, we introduce a bidirectional merging approach to merge the mask predictions from different frames, as shown in \figref{fig:bmm}.

Specifically, we denote the point cloud from two adjacent frames as $X^1=\{x^1_1, x^1_2, \cdots, x^1_m\}$ and $X^2=\{x^2_1, x^2_2, \cdots, x^2_n\}$, where $m$ and $n$ are the numbers of points in $X^1$ and $X^2$, respectively. 
We compute the correspondence mapping $M$ between $X^1$ and $X^2$.
The points $x_{i}^1$ and $x_{j}^2$ are a pair of matched points between $X^1$ and $X^2$ if $(i,j)\in M$.
For each mask id $m$ in $X^1$, we first find the points in $X^1$ within the same 3D mask, denoted as $Q^1$. The number of points with mask id $m$ in $X^1$ is denoted as $\sigma^1_m$. We then leverage the correspondence mapping $M$ to find their corresponding points in $X^2$, denoted as $Q^2$.
For each mask id $n$ of the points in $Q^2$, we denote $\sigma^{1-2}_{mn}$ as the number of points in $Q^2$ whose mask id is $n$. 
Similarly, we denote $\sigma^2_n$ as the number of points in $X^2$ whose mask id is $n$.
If $\sigma^{1-2}_{mn} > \delta \times min(\sigma_{m}^{1}, \sigma_{n}^{2})$ (where $\delta$ is the threshold and it is set to $0.5$ in our experiments), it means that the mask with id $m$ from $X^1$ and the mask with id $n$ from $X^2$ are highly overlapping masks. So we merge the masks by assigning $n$ as the new mask id for points in $X^2$ whose previous mask ids are $m$.
This process is followed by a symmetric process where we iterate the mask ids in $X^2$ and query the points and mask ids in $X^1$.

\subsection{Bottom-up Merging of the Point Clouds in the Whole Scene}
\label{sec:rmm}
In ScanNet~\cite{dai2017scannet}, a complex scene consists of hundreds of or thousands of frames where each frame corresponds to a local point cloud.
We propose a bottom-up merge approach to merge the masks of local point clouds into the consistent masks of the whole scene, as shown in \figref{fig:rmm}.

Let the local point clouds be $X_0^1, X_0^2, \cdots, X_0^K$ where $K$ is the total number of frames in the scene.
We first merge point clouds from adjacent frames $X_0^{2i}$ and $X_0^{2i+1}$ into a new point cloud $X_1^i$.
Then based on the new point clouds $X_1^1, X_1^2, \cdots, X_1^{K/2}$, we merge $X_1^{2i}$ and $X_1^{2i+1}$ into $X_2^i$.
We repeat this process for $\ceil{\log_2 n}$ times until the point clouds are merged into a single point cloud $X_{\ceil{\log_2 n}}^0$ of the whole scene.
We leverage the aforementioned bidirectional merging approach for each step of merging between two adjacent point clouds, and apply grid pooling~\cite{wu2022point} to downsample the point clouds after each merging step.
Mathematically, we denote $\oplus$ as the bidirectional merging operation, and the bottom-up merging rule can be formulated as,
\begin{equation}
X_{t+1}^{i} = X_t^{2i}\oplus X_t^{2i+1},
\end{equation}
where the two adjacent frames $X_t^{2i}$ and $X_t^{2i+1}$ at step $t$ are merged into one frame $X_{t+1}^{i}$ at step $t+1$.

\subsection{Mask Ensemble with Over-segmentation}
\label{sec:sre}

Following ScanNet~\cite{dai2017scannet}, we obtain the over-segmentation masks of the 3D scenes by applying a normal-based graph cut method~\cite{felzenszwalb2004efficient,karpathy2013object}, which takes the geometric information into consideration.
In contrast, our SAM3D approach is based on Segment Anything Model~\cite{kirillov2023segany} from the semantics and edges of the RGB frames.
To obtain higher-quality masks, we further merge the point cloud masks generated by our SAM3D with the over-segmentation results with the bidirectional merging approach.
The ensembled masks are generated with both 3D geometric information and RGB semantics and edges information, which are shown to demonstrate higher segmentation quality.

%% file: tex/4_experiments.tex
\section{Results}
The segmentation results of ScanNet~\cite{dai2017scannet} 3D scenes are shown in \figref{fig:results}. Our framework can achieve accurate and fine-grained segmentation results. Our approach can even generate more detailed masks than the ground-truth annotations. For example, in the first row of \figref{fig:results}, the paintings or pictures on the wall are not annotated in the ground-truth masks, and not segmented by the over-segmentation. However, our SAM3D approach shown in the second column is able to segment the paintings on the wall, such as the category of \textit{picture}.

%% file: tex/5_conclusion.tex
\section{Conclusion and Discussion}
In summary, we propose a novel framework SAM3D for generating the segmentation masks of 3D scenes. The proposed approach projects the 2D masks predicted by the Segment-Anything Model (SAM) to 3D point clouds, and further merges the segmentation masks of different point cloud frames to obtain the segmentation masks of the whole 3D scene. Our proposed SAM3D obtains accurate and fine-grained masks for the 3D scenes.

Our proposed SAM3D proves that the segmentation results by SAM can provide a strong baseline and initial state for 3D segmentation. We will further investigate its applications in open-vocabulary 3D segmentation and segment anything in 3D.